# UNIVERSAL GENERATIVE MODELING FOR CALIBRATION-FREE PARALLEL MR IMAGING

*Wanqing Zhu[1], Bing Guan[1], Shanshan Wang[2], Minghui Zhang[1] and Qiegen Liu[1]*

[1]Department of Electronic Information Engineering, Nanchang University, Nanchang 330031, China

[2]Paul C. Lauterbur Research Center for Biomedical Imaging, SIAT, CAS, Shenzhen 518055, China

## ABSTRACT

The integration of compressed sensing and parallel imaging (CS-PI) provides a robust mechanism for accelerating MRI acquisitions. However, most such strategies require the explicit formation of either coil sensitivity profiles or a cross-coil correlation operator, and as a result reconstruction corresponds to solving a challenging bilinear optimization problem. In this work, we present an unsupervised deep learning framework for calibration-free parallel MRI, coined universal generative modeling for parallel imaging (UGM-PI). More precisely, we make use of the merits of both wavelet transform and the adaptive iteration strategy in a unified framework. We train a powerful noise conditional score network by forming wavelet tensor as the network input at the training phase. Experimental results on both physical phantom and in vivo datasets implied that the proposed method is comparable and even superior to state-of-the-art CS-PI reconstruction approaches.

## 1. INTRODUCTION

Magnetic Resonance Imaging (MRI) plays an indispensable role in many clinical and research scenarios. Unfortunately, the inherently data acquisition properties make it slower than modalities like X-ray, Computed tomography, or Ultrasound. To alleviate this issue without degrading image quality significantly, many efficient acceleration schemes for MRI acquisition have been studied, which can be mainly divided into two aspects: The compressed sensing MRI (CS-MRI) [1], [2] and the parallel imaging (PI) [3], [4]. These methods are designed to facilitate accurate reconstruction from the insufficient k-space samples.

For CS-MRI, it can speed up MRI acquisition by acquiring fewer samples than required by utilizing the sparsity of the data in the transformed domain. Specifically, image sparsity in a transform domain is incorporated into the classic formulation of compressed sensing in MRI [5] utilizes as the prior information. However, the time-consuming iterations and the challenge to choose the optimal transform hinder its wide application. For PI, it uses multiple channels for simultaneous data acquisition. Typical examples include sensitivity encoding (SENSE) [3], generalized auto-calibrating partially parallel acquisitions (GRAPPA) [4], iterative self-consistent parallel imaging reconstruction (SPIRiT) [6], and so on [7]-[10]. In recent years, the compressed sensing based parallel imaging (CS-PI) technique has shown an increased popularity. Based on how to use the sensitivity information, these techniques can be roughly categorized into pre-calibrated, auto-calibration, and calibration-free approaches. Both of the former two kinds need to estimate the sensitivity information from a separate calibration scan or auto-calibration signal (ACS). In contrast, calibration-free techniques don't need specific calibration regions/pre-scans to estimate the interpolation kernels or sensitivities [11]. Typical examples include simultaneous auto-calibrating and k-space estimation (SAKE) [12] and P-LORAKS [13].

Recently, the approaches that learned image priors through deep learning have rapidly gained in popularity [14]-[17]. There are mainly two categories: Supervised and unsupervised schemes [18]. Supervised learning utilizes data information to train a universal network for learning the mapping between the under-sampled and fully sampled data pairs, but the training of supervised learning requires a huge number of labeled samples to ensure proper convergence. Unsupervised learning systems such as RAKI and PixelCNN [19], [20] unroll the optimization algorithm iterations to the neural network so that the network can automatically learn the hyperparameters or transformations in the optimization algorithm. Nevertheless, there are limitations of adding a layer of complexity when dealing with multi-coil data [21], and unsupervised deep learning for parallel imaging (DL-PI) is less explored in the existing research.

To solve the above issues, we propose a more flexible and more universal unsupervised deep learning framework for calibration-free parallel MRI in this paper. At first, different from the existing networks for parallel imaging, we can apply the prior of single-coil image in wavelet transform domain to parallel imaging scenes with any number of coils. Additionally, based on the core work in method HGGDP [22], we propose a more efficient adaptive iteration strategy for reducing the iteration number of the inner loop. Constructing prior information in wavelet higher-dimensional space alleviates the issues of low-dimensional manifold and low data density region in generative models. Moreover, the proposed method could exploit redundant characteristics of the wavelet domain implicitly. The main contributions of this work are summarized as follows:

- **Universal generative modeling**: A multi-channel and multi-scale tensor that formed by undecimated wavelet transform serves as the generative network input. Although the learned prior knowledge is trained from single coil image, it can be used for PI reconstruction scenarios with any coil number. The proposed model has strong flexibility and robustness.
- **Reconstruction with fast mixing**: Due to the special similarity among the multi-channel object, a more efficient adaptive iteration strategy for reducing the iteration number of the inner loop is introduced to replace the original fixed number of iterations. Thus, the whole calculation amount is reduced.

This work was supported by National Natural Science Foundation of China (61871206, 61601450).

## 2. PRELIMINARIES

### 2.1. CS-PI

Under-sampled data acquisition reconstructed by CS-PI can potentially reduce trial cost and improve compliance. Let $M \in \mathbb{C}^{N \times N}$ and $F \in \mathbb{C}^{N \times N}$ respectively denote the mask indicating the under-sampling k-space locations and the normalized full Fourier encoding matrix, i.e., $F^H F = I_N$. The k-space data acquisition for multi-coil parallel MRI can be described as follows:

$$y_j = F_M x_j + \eta_j, \quad j = 1, 2, \cdots, J \quad (1)$$

where $x_j \in \mathbb{C}^N$ represents the vectorized $j$-th coil image, $F_M = MF \in \mathbb{C}^{N \times N}$, $\eta_j$ is the noise, and $J$ is the total number of receiver coils.

To estimate the reconstructed image of $J$ coils, we can minimize the following unconstrained optimization problem:

$$X = \arg\min_x \left\{ \frac{1}{2} \sum_{j=1}^{J} \| F_M x_j - y_j \|_2^2 + \lambda \psi(x_j) \right\} \quad (2)$$

where $X = [x_1, x_2, \cdots, x_J] \in \mathbb{C}^{N \times J}$ stacks the vectors as columns, $\lambda$ is the regularization parameter that balances the two terms and $\psi(x_j)$ represents the sparsity regularization term. The final image is obtained from the square root of the sum of squares (SOS) image.

### 2.2. Deep gradient of prior in wavelet domain

In recent years, most existing generative models are devoted to describing the data distribution faithfully [23], [24]. Nevertheless, it will cause the problem of inaccurate representation of the samples. In order to solve the sample representation problem, instead of estimating data density $p_{data}(x)$, Song *et al.* presented a new generative model named noise conditional score networks (NCSN) that trains a score network $S_\theta(x, \sigma)$ parameterized by $\theta$, which directly estimate the gradient of data density $\nabla_x \log p_\sigma(x)$ [25]. The network employs annealed Langevin dynamics for image generation (i.e., $p_\sigma(x) \approx p_{data}(x)$, $\sigma \to 0$) under a sequence of noise levels $\{\sigma_i\}_{i=1}^{I}$. Supposing the noise distribution is selected as $p_\sigma(\tilde{x} | x) = N(\tilde{x} | x, \sigma^2 I)$, it subsequently leads to $\nabla_{\tilde{x}} \log p_\sigma(\tilde{x} | x) = -(\tilde{x} - x) / \sigma^2$. For a given $\sigma$, the objective of NCSN is:

$$\ell(\theta; \sigma) \triangleq \frac{1}{2} E_{p_\sigma(\tilde{x}, x)} [\| S_\theta(\tilde{x}, \sigma) + (\tilde{x} - x) / \sigma^2 \|_2^2] + C \quad (3)$$

Although NCSN has achieved good performances, it suffers from two major deficiencies: low data density regions and the manifold hypothesis [26], [27]. In order to circumvent these restrictions, we propose a universal generative model (UGM-PI) to enrich the native NCSN by the usage of the deep gradient of prior in higher-dimensional wavelet space. More specifically, we utilize wavelet transformation $X = T_w(x)$ to establish a 4-channel higher-dimensional tensor. Then, the components are as follows:

$$X = \{x_{ll}, x_{lh}, x_{hl}, x_{hh}\} \quad (4)$$

Subsequently, instead of estimating low data density regions $\nabla_x \log p_\sigma(x)$, the present UGM-PI is trained with $X$ at high-dimensional manifold and high-density regions as network input. More precisely, we generate a list of noise levels $\{\sigma_i\}_{i=1}^{I}$ that are reduced proportionally for each step of the outer loop. Artificial Gaussian noise is added to the iteration stage in the light of the noise level $\sigma_i$ from high to low. Then, Eq. (3) can be rewritten as:

$$\begin{aligned}
\ell(\theta; \sigma) &\triangleq \frac{1}{2} E_{p_\sigma(X)} [\| S_\theta(X, \sigma) - \nabla_x \log p_\sigma(X) \|_2^2] \\
&= \frac{1}{2} E_{p_\sigma(\tilde{X}, X)} [\| S_\theta(\tilde{X}, \sigma) - \nabla_{\tilde{X}} \log p_\sigma(\tilde{X} | X) \|_2^2] + C \quad (5) \\
&= \frac{1}{2} E_{p_\sigma(\tilde{X}, X)} [\| S_\theta(\tilde{X}, \sigma) + (\tilde{X} - X) / \sigma^2 \|_2^2] + C
\end{aligned}$$

where $C$ is a constant that does not depend on $\theta$.

After Eq. (5) is derived, the loss function is combined with all $\sigma \in \{\sigma_i\}_{i=1}^{I}$ to get one unified objective:

$$L(\theta; \{\sigma_i\}_{i=1}^{I}) \triangleq \frac{1}{I} \sum_{i=1}^{I} \lambda(\sigma_i) \ell(\theta; \sigma_i) \quad (6)$$

where $\lambda(\sigma_i) > 0$ refers to a coefficient function depending on $\sigma_i$. Since Eq. (6) is a conical combination of $I$ and UGM-PI objectives, the optimal score $S_{\theta^*}(X, \sigma)$ minimizes Eq. (6) while $S_{\theta^*}(X, \sigma_i) = \nabla_X \log p_{\sigma_i}(X)$ is satisfied for all $i \in \{1, 2, \cdots, I\}$.

The benefit of constructing deep learning priors in wavelet domain is visualized in Fig. 1. It depicts the convergence tendency of PSNR curves versus iteration for a 12 coils brain image that reconstructed by UGM-PI and the native NCSN, respectively. Even if the final reconstruction result is near, the introduction of the deep gradient of prior in higher-dimensional wavelet space enables the learning model to reach the convergence target more quickly. The multi-scale wavelet decomposition subdivides the original problem into a group of smaller subproblems, so that the generative model can better learn the gradient of data density, which is the key to faster convergence.

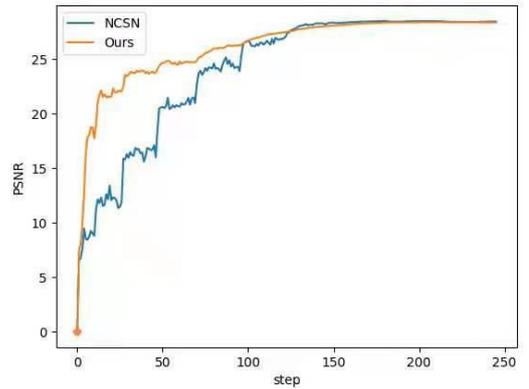

**Fig. 1**. Convergence tendency comparison of DSM in the native NCSN and the advanced UGM-PI, respectively.

### 2.3. Reconstruction with universal deep prior

For the neural network trained by high-dimensional wavelet tensor $X$, it satisfies $S_\theta(X, \sigma_i) \approx \nabla_x \log p_{\sigma_i}(x)$ for all $i \in \{1, 2, \cdots, I\}$. By means of the artificial noise method, the reconstruction result is obtained by an operation that makes noisy high-dimensional tensor as the input of $S_\theta(X, \sigma)$ via gradually annealed noise. i.e.,

$$X^t = X^{t-1} + \frac{\alpha_i}{2}\nabla_x \log p_{\sigma_i}(X^{t-1}) + \sqrt{\alpha_i}z_t$$
$$= X^{t-1} + \frac{\alpha_i}{2}S_\theta(X^{t-1},\sigma_i) + \sqrt{\alpha_i}z_t \quad (7)$$

where $\alpha_i$ represents the step size by tuning down it gradually. The initial solution $X^0$ can be a total uniform noise or other pre-defined value. The index set $t=1,\cdots,T$ is defined at the inner loop of each $\sigma_i$, while the index set $i=1,\cdots,I$ goes over the outer loop of every $\sigma_i$.

For the iteration number $T$ of the inner loop, it should be emphasized that an adaptive iteration strategy is proposed for reducing the calculation amount. Since the level of Gaussian noise added in the reconstruction stage decreases from high to low, it only provides an initial value while the added noise level is high. Therefore, too many iterations are of little significance for model optimization. In order to improve the reconstruction efficiency of the model, a stepwise reduction method is developed. Specifically, we set the inner loop number of the $\sigma_i$ with the function $10(\log_e i + 1)$. This strategy not only reduces the computational cost, but also maintains the reconstruction performance.

Besides, we add a data consistency during the reconstruction process. Specifically, at each iteration of the annealed Langevin dynamics, we update the solution via data consistency constraint after Eq. (7), i.e., let $x^t = T_w^{-1}(X^t)$, it yields,

$$x_j^{t+1} = \arg\min_x \left\{ \sum_{j=1}^J \|F_m x_j - y_j\|_2^2 + \lambda \|x_j - x_j^t\|_2^2 \right\} \quad (8)$$

Here, we empirically set $\lambda=1$ and it works well in our experiments.

The least-square (LS) minimization in Eq. (8) can be solved as follows:

$$(F_m^T F_m + \lambda)x_j^{t+1} = F_m^T y_j + \lambda x_j^t \quad (9)$$

Let $F \in C^{M \times M}$ denotes the full Fourier encoding matrix which is normalized as $F^T F = 1_M$. $Fx_j(k_v)$ stands for the updated $j$-th coil value at under-sampled k-space location $k_v$, and $\Omega$ represents the sampled subset of data, it yields,

$$Fx_j(k_v) = \begin{cases} Fx_j^t(k_v), & k_v \notin \Omega \\ \dfrac{FF_m^T y_j(k_v) + \lambda Fx_j^t(k_v)}{(1+\lambda)}, & k_v \in \Omega \end{cases} \quad (10)$$

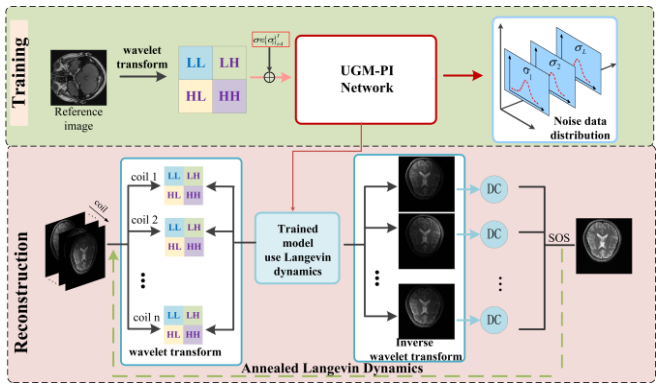

**Fig. 2.** Flowchart illustration of the proposed UGM-PI model. Top: Network training process in single coil image for prior learning. Bottom: An intermediate result of the iterative process at multi-coil MRI reconstruction phase.

Then, the $j$-th coil image $x_j$ is obtained by the inverse Fourier transform of the frequency data in the above equation, and the final image is produced as the square root of the SOS image. A visual illustration of the proposed UGM-PI model is depicted in Fig. 2. In summary, the main reconstruction steps of UGM-PI are listed as follows:

| Algorithm 1 UGM-PI |
|---|
| **Training stage** |
| **Dataset**: Dataset in wavelet domain: $X=\{x_{ll},x_{lh},x_{hl},x_{hh}\}$ |
| **Output**: Trained UGM-PI $S_\theta(X,\sigma)$ |
| **Reconstruction stage** |
| **Setting:** $\sigma \in \{\sigma_i\}_{i=1}^I$, $\varepsilon, T$, $x^0$, $k_v$ and $\Omega$ |
| 1: **for** $i \leftarrow 1$ to $I$ **do** (Outer loop) |
| 2: $\quad \alpha_i = \varepsilon \cdot \sigma_i^2 / \sigma_I^2$ |
| 3: $\quad$ **for** $t \leftarrow 1$ to $T$ **do** (Inner loop) |
| 4: $\quad\quad X_j^t = T_W(x_j^t)$ |
| 5: $\quad\quad$ Draw $z_t \sim N(0,1)$ and $X^{t-1}=\{x_{ll}^{t-1},x_{lh}^{t-1},x_{hl}^{t-1},x_{hh}^{t-1}\}$ |
| 6: $\quad\quad X_j^t = X_j^{t-1} + \frac{\alpha_i}{2}S_\theta(X_j^{t-1},\sigma_i) + \sqrt{\alpha_i}z_t$ |
| 7: $\quad\quad$ Update $x_j^t = T_W^{-1}(X_j^t)$ and Eq. (10) |
| 8: $\quad$ **end for** |
| 9: $\quad x_j^0 \leftarrow x_j^T$ |
| 10: $\quad$ Update multi-coil images $x_j^T$, $j=1,\cdots,J$ |
| 11: **end for** |
| 12: Update the final image as the square root of $SOS(x_j^T)$ |

## 3. EXPERIMENTS

### 3.1. Experiment setup

**Datasets.** At the training phase, we use brain images from *SIAT* dataset, which was provided by Shenzhen Institutes of Advanced Technology, the Chinese Academy of Science.scanned. Informed consents were obtained from the imaging subject in compliance with the institutional review board policy. All data were scanned from a 3T Siemens's MAGNETOM Trio scanner using the T2-weighted turbo spin echo sequence. The TE time was 149 ms, TR was 2500 ms, the field of view (FOV) was 220×220 mm$^2$ and the slice thickness was 0.86 mm. In addition, the number of receiver coils is 12 and the collected dataset includes 500 2D complex-valued MR images. At the reconstruction stage, the experiments were conducted on four datasets [28], which contain physical phantom and in vivo datasets.

**Sampling masks**. The under-sampled operation in the Fourier domain is realized by using 2D random under-sampling with the variable density pattern, and 2D Poisson disk under-sampling. For the reconstruction comparison on different sampling rate, the accelerated factors are varied over three values, $R=$ 4, 6, and 10.

**Compared methods**. To evaluate the proposed method, it is compared against three state-of-the-art calibration-free methods, namely P-LORAKS, SAKE, and learn joint-sparse codes for calibration-free parallel MR imaging (LINDBREG) [28]. The first two methods belong to k-space-based approach, the latter method and proposed method belong to image-based approach. For the convenience of reproducible research, source code of UGM-PI can be downloaded from website: *https://github.com/yqx7150/UGM-PI*.

### 3.2. Comparison results with 2D under-sampling scheme

The present UGM-PI is compared with the competing methods under two sampling schemes, namely 2D variable density random and 2D variable density Poisson masks. These under-sampling schemes tend to produce noise-like artifacts during the reconstruction process. It is worth noting that the test images are multi-coil MRI data, and UGM-PI model is still trained on single-coil dataset. Fig. 3 presents the physical phantom reconstructed by these methods under 2D variable density random mask with the acceleration factor $R$= 6 and the corresponding reconstruction error maps. Conclusion can be reached from the reconstruction results and error maps that UGM-PI exhibits the least error.

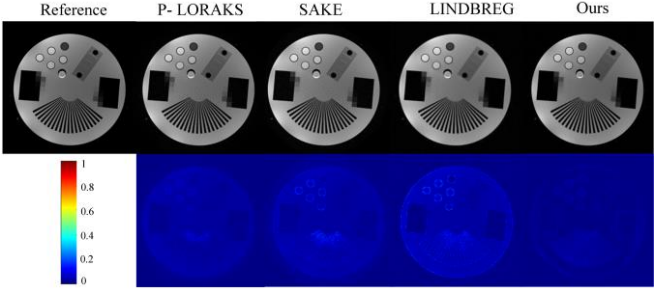

**Fig. 3.** The test comparison of 4 coils physical phantom obtained by P-LORAKS, SAKE, LINDBREG, and the proposed UGM-PI under 2D random sampling with the acceleration factor $R$=6.

For more comparisons, visualization results reconstructed by different methods under 2D variable density Poisson sampling trajectory with the acceleration factor $R$= 10 are provided in Fig. 4. The corresponding reconstruction error has also been presented for better comparison purpose. It can be observed that SAKE and LINDBREG still suffer from artifact, while P-LORAKS and UGM-PI produce better reconstruction results. As for P-LORAKS, the corresponding reconstruction errors show that artifact noise becomes less severe compared to the other results but still exists. UGM-PI incorporates sparsity constraint, which is beneficial to reducing the noisy issue to some extent.

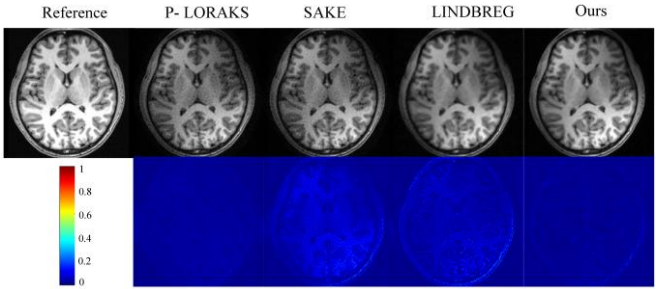

**Fig. 4.** Reconstruction comparison of the in vivo MPRAGE brain image obtained by P-LORAKS, SAKE, LINDBREG, and UGM-PI under 2D variable density Poisson sampling with the acceleration factor $R$=10.

In addition to the visual comparison, we further provide the average quantitative measurement comparisons for these methods with respect to different acceleration factors. All methods are evaluated to reconstruct both 4 coils physical phantom and in-vivo datasets with two kinds of 2D sparse sampling schemes at high acceleration factors. The quantitative averaged results were summarized in Table I. We can observe that the proposed method has the highest PSNR values than P-LORAKS, SAKE, and LINDBREG under both two kinds of sampling patterns. When the acceleration factor increases to be only 10% sampling rate, the reconstruction result gradually deteriorates. Nevertheless, even at 10x acceleration under 2D Poisson disk sampling, the reconstruction performance obtained by UGM-PI is still acceptable. To sum up, UGM-PI can achieve more satisfactory results with clearer contours, sharper edges, and finer image details under various sampling masks.

TABLE I
AVERAGE PSNR, SSIM AND HFEN VALUES OF RECONSTRUCTION RESULTS BY DIFFERENT ALGORITHMS UNDER THE 2D VARIABLE DENSITY SAMPLING AND DIFFERENT SAMPLING TRAJECTORIES WITH THE SAME PERCENTAGE.

| | | P-LORAKS | SAKE | LIND-BREG | UGM-PI |
|---|---|---|---|---|---|
| Random _2D | $R$=4 | 35.94<br>0.92<br>0.50 | 35.95<br>0.88<br>0.57 | 33.49<br>0.93<br>0.63 | **40.13**<br>**0.95**<br>**0.40** |
| | $R$=6 | 32.73<br>0.88<br>0.65 | 33.62<br>0.84<br>0.80 | 30.79<br>0.90<br>0.80 | **37.25**<br>**0.93**<br>**0.53** |
| Poisson _2D | $R$=6 | 33.61<br>0.90<br>0.48 | 34.96<br>0.87<br>0.57 | 32.47<br>0.92<br>0.59 | **38.05**<br>**0.95**<br>**0.42** |
| | $R$=10 | 31.22<br>0.85<br>0.83 | 32.13<br>0.82<br>0.94 | 28.67<br>0.84<br>1.19 | **34.20**<br>**0.91**<br>**0.71** |

In order to intuitively describe the convergence process of the model, we plot the convergence tendency of the PSNR curve with iteration during reconstructing the 12 coils brain image under 2D randomly under-sampling mask with the acceleration factor $R$= 4 in Fig. 5. As can be seen, the final reconstructed result is evolved from the pure random noise. In addition, the whole reconstruction process is convergent, which provides a foundation for us to propose an adaptive iteration strategy that reduces iteration number of the inner loop.

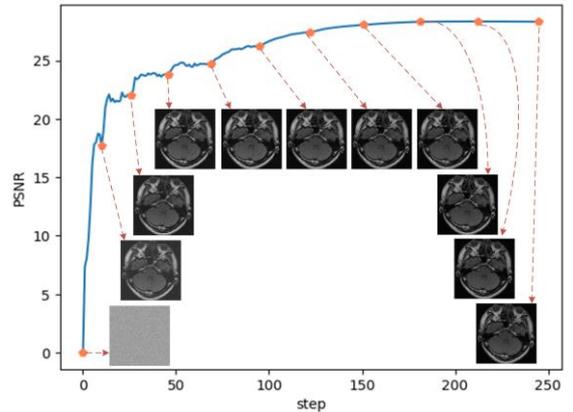

**Fig. 5.** Visualization of the intermediate convergence process of UGM-PI.

## 4. CONCLUSIONS

In this study, a universal generative modeling for calibration-free parallel MR imaging reconstruction termed UGM-PI was presented. Specifically, we made use of the merits of wavelet transform at the prior learning stage and the adaptive iteration strategy at the reconstruction stage. The proposed method has been compared to three representative calibration-free approaches. Experimental results demonstrated that the proposed UGM-PI can produce images with less noise and artifacts than the state-of-the-art methods. In the forthcoming future, we may investigate other more special-designed transforms for promoting the accuracy of the parallel imaging reconstruction.